\documentclass[10pt,twocolumn,letterpaper]{article}

\usepackage{cvpr}              
\definecolor{cvprblue}{rgb}{0.21,0.49,0.74}
\usepackage[pagebackref,breaklinks,colorlinks,allcolors=cvprblue]{hyperref}

\def\paperID{11244} 
\def\confName{CVPR}
\def\confYear{2026}

\title{SEATrack: Simple, Efficient, and Adaptive Multimodal Tracker}

\author{Junbin Su\textsuperscript{1$\ast$}, Ziteng Xue\textsuperscript{2$\ast$}, Shihui Zhang\textsuperscript{1,3\Letter}, Kun Chen\textsuperscript{1}, Weiming Hu\textsuperscript{4,5}, Zhipeng Zhang\textsuperscript{6\Letter}
\\
\small \textsuperscript{1}School of Artificial Intelligence (School of Software), Yanshan University\qquad
\small \textsuperscript{2}School of Software, Beihang University\\
\small \textsuperscript{3}Hebei Key Laboratory of Computer Virtual Technology and System Integration\\
\small \textsuperscript{4}State Key Laboratory of Multimodal Artificial Intelligence Systems (MAIS), CASIA\qquad
\small \textsuperscript{5}School of Artificial Intelligence, UCAS\\
\small \textsuperscript{6}AutoLab, School of Artificial Intelligence, Shanghai Jiao Tong University
}
\usepackage{bibentry}
\usepackage{marvosym}
\usepackage{fontawesome5}
\usepackage{amsmath}
\usepackage{amssymb}
\usepackage{xcolor}
\usepackage{booktabs}
\usepackage{colortbl}
\usepackage{latexsym}
\usepackage{pifont}
\usepackage{multirow,array,makecell}

\begin{document}
\maketitle
\footnotetext{$^\ast$ Equal contribution\quad\textsuperscript{\Letter} Corresponding author}
\begin{abstract}
Parameter-efficient fine-tuning (PEFT) in multimodal tracking reveals a concerning trend where recent performance gains are often achieved at the cost of inflated parameter budgets, which fundamentally erodes PEFT's efficiency promise. In this work, we introduce SEATrack, a Simple, Efficient, and Adaptive two-stream multimodal tracker that tackles this performance-efficiency dilemma from two complementary perspectives. We first prioritize cross-modal alignment of matching responses, an underexplored yet pivotal factor that we argue is essential for breaking the trade-off. Specifically, we observe that modality-specific biases in existing two-stream methods generate conflicting matching attention maps, thereby hindering effective joint representation learning. To mitigate this, we propose AMG-LoRA, which seamlessly integrates Low-Rank Adaptation (LoRA) for domain adaptation with Adaptive Mutual Guidance (AMG) to dynamically refine and align attention maps across modalities. We then depart from conventional local fusion approaches by introducing a Hierarchical Mixture of Experts (HMoE) that enables efficient global relation modeling, effectively balancing expressiveness and computational efficiency in cross-modal fusion. Equipped with these innovations, SEATrack advances notable progress over state-of-the-art methods in balancing performance with efficiency across RGB-T, RGB-D, and RGB-E tracking tasks. \href{https://github.com/AutoLab-SAI-SJTU/SEATrack}{\textcolor{cyan}{Code is available}}.

\end{abstract}    
\section{Introduction}
\label{sec:intro}

Object tracking aims to localize a target in a video given its initial appearance. While modern RGB trackers handle most scenarios effectively, their reliance on a single modality renders them vulnerable to common real-world challenges like drastic illumination changes and rapid motion. Consequently, multimodal tracking has gained substantial traction by fusing complementary data sources to achieve the robustness required to handle such difficult cases.

\begin{figure}[t]
    \centering
    \includegraphics[width=\linewidth]{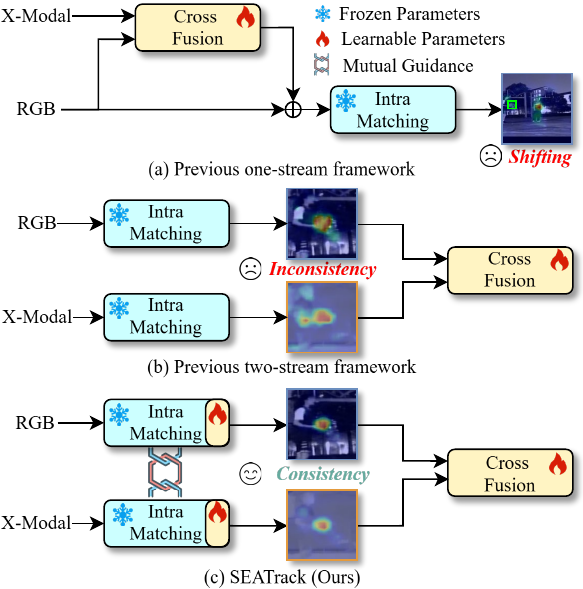}
        \caption{\textbf{Previous frameworks \textit{v.s.} SEATrack}. (a) The previous one-stream method~\cite{zhu2023visual} suffers from attention shifting when performing intra-modal matching on mixed inputs. (b) Similarly, domain gaps cause attention maps' inconsistency in the two-stream method~\cite{hou2024sdstrack}. (c) Our method is able to produce aligned and refined attention maps, which facilitate cross-modal fusion.} 
    \label{fig:abs}
    \vspace{-10pt}
\end{figure}

A prevailing strategy in multimodal tracking is to adapt pre-trained RGB trackers for heterogeneous modalities such as thermal, depth, or event data (collectively, X-modal). While leveraging powerful prior knowledge, the approach faces notable limitations under both full fine-tuning (FFT) and parameter-efficient fine-tuning (PEFT) paradigms. Specifically, with FFT, updating the entire network with limited multimodal data often leads to catastrophic forgetting of learned knowledge~\cite{he2021towards, jia2022visual, huang2023vop}. Furthermore, the substantial computational cost of FFT makes it impractical~\cite{hui2023bridging, chen2024simplifying, hu2025exploiting}. These drawbacks motivate a focus on the PEFT paradigm. Yet, a concerning trend also emerges. To achieve or approach the performance of FFT, recent state-of-the-art (SOTA) PEFT methods~\cite{hou2024sdstrack, wu2024single, hong2024onetracker,tan2024xtrackmultimodaltrainingboosts} have drastically increased the number of tunable parameters, even up to 16 times more than the pioneering work~\cite{zhu2023visual}. This trajectory arguably contradicts the core principle of efficiency that justifies PEFT in the first place.

Beyond the inefficiencies in parameters, significant limitations are also present in the cross-modal alignment of existing PEFT-based trackers. As common sense, a crucial function in tracking is the matching of features between the template and the search region, which is usually accomplished via self-attention in recent SOTA approaches~\cite{ye2022joint, gao2023generalized, lin2024tracking, zheng2024odtrack, shi2024explicit}. Multimodal trackers inherit this design for intra-modal matching but with suboptimal outcomes. In one-stream architectures, matching is performed on mixed-modality features. However, the knowledge encapsulated in the pre-trained model is intrinsically biased towards the RGB domain. When these model weights are kept fixed during training, this domain-specific bias prevents effective generalization to the mixed distribution, manifesting as erroneous attention shifts (Fig.\ref{fig:abs}(a)). This phenomenon reinforces the conclusion from~\cite{hou2024sdstrack} that one-stream designs are less robust than two-stream methods. For two-stream ones, this domain gap persists and is compounded by dynamic modality reliability, leading to inconsistent matching in which conflicting attention maps from different modalities hinder effective joint representation learning (Fig.\ref{fig:abs}(b)). Besides, a separate long-standing challenge lies in designing cross-modal fusion strategies that balance expressiveness and efficiency. Although powerful, attention-based fusion mechanisms incur significant computational overhead due to their quadratic complexity~\cite{hui2023bridging,chen2024simplifying,hou2024sdstrack,xiao2025cross}. In contrast, lightweight alternatives like local fusion~\cite{zhu2023visual,wu2024single,hong2024onetracker, cao2024bi,tan2024xtrackmultimodaltrainingboosts} lack a global respective field, which limits their expressive power.

This makes us wonder: \textit{is there a way to balance performance and efficiency in multimodal tracking elegantly?} We provide an affirmative answer to this challenge by proposing SEATrack, a multimodal tracker designed to be \textbf{S}imple, \textbf{E}fficient, and \textbf{A}daptive. Architecturally, it employs a two-stream design for tracking robustness and a PEFT paradigm for training efficiency.

To address the critical challenges of domain gap and attention inconsistency within existing two-stream models, we introduce Adaptive Mutual Guidance Low-Rank Adaptation (AMG-LoRA). The central idea of our approach stems from a fundamental hypothesis that, since multimodal inputs are spatio-temporally aligned, the responses of intra-modal target matching should, in principle, be consistent with each other. We leverage this expected consistency by proposing a cross-modal guidance mechanism. Specifically, the matching information derived from one modality is used to condition and guide the matching process in the other. This reciprocal exchange enables mutual reinforcement between modalities, yielding more robust joint representations. More concretely, we first employ LoRA within the attention layer's projection matrices to facilitate domain adaptation. Recognizing that the target's selective emphasis on each modality varies with the scenario~\cite{cao2024bi, hu2024towards}, we contend that the alignment must be dynamic. We achieve this through Adaptive Mutual Guidance (AMG), a mechanism that dynamically refines and aligns attention maps across modalities. Inspired by classifier-free guidance (CFG)~\cite{ho2022classifier}, AMG reformulates cross-modal alignment as a multi-branch trade-off problem. It treats one modality's discriminative prior as the ``unconditional'' branch and the other as the ``conditional'' branch, enabling bidirectional interaction. As we will show, AMG-LoRA effectively addresses domain gaps and attention alignment (Fig.\ref{fig:abs}(c)) while boosting cross-modal fusion with negligible additional latency. In response to the long-standing challenge in cross-modal fusion, we introduce an efficient global relation modeler dubbed Hierarchical Mixture-of-Experts (HMoE). Distinct from existing explorations~\cite{chen2024emoetrackerenvironmentalmoebasedtransformer, cai2025spmtrack,tan2025xtrack} that focus on expert-level ensemble learning via top-$k$ or dense routing, HMoE further enables fine-grained interaction from sub-token to token level through a hierarchical soft routing. In this way, HMoE functions as an effective token mixer for cross-modal fusion. Empirically, HMoE achieves about 35\% faster speed (FPS) compared to its attention-based counterpart, while maintaining comparable performance.


Overall, our contributions are: $\spadesuit$ We introduce SEATrack, a Simple, Efficient, and Adaptive multimodal tracker that achieves a new state-of-the-art result in balancing performance with efficiency. \ding{70} We target the underexplored challenge of cross-modal attention alignment by proposing Adaptive Mutual Guidance Low-Rank Adaptation (AMG-LoRA). Extensive experiments show that it holds strong potential as a solid baseline, offering a novel perspective on resolving the performance-efficiency dilemma. $\clubsuit$ We propose Hierarchical MoE (HMoE) as the response to the long-standing challenge in cross-modal fusion. As an efficient global relation modeler, it delivers about a 35\% FPS speed-up over attention-based counterparts while maintaining comparable performance.
\vspace{-10pt}
\section{Related Works}
\label{sec:RW}

\subsection{Multimodal Tracking}
While RGB-only tracking achieves remarkable success, it still struggles in challenging scenarios. Therefore, several multimodal datasets~\cite{li2019rgb,li2021lasher,yan2021depthtrack} and methods~\cite{qian2021dal,gao2019deep,zhang2021jointly} have been proposed to enable all-weather and all-day tracking. Existing methods can be roughly categorized into two branches based on fine-tuning strategies. The first branch fully fine-tunes (FFT) pre-trained RGB trackers (termed foundation tracker) to different fields. Specifically, DeT~\cite{yan2021depthtrack} attaches the depth flow branch to a ResNet-based foundation tracker and fine-tunes it on RGB-D datasets. TBSI~\cite{hui2023bridging} pioneers the extension of ViT-based foundation trackers for RGB-T tracking. However, as foundation trackers evolve with larger backbones, FFT becomes both computationally prohibitive and prone to catastrophic forgetting on limited multimodal data~\cite{he2021towards,jia2022visual, huang2023vop}.

The aforementioned drawbacks motivate a focus on the PEFT paradigm, which freezes all or most of the parameters of the foundation tracker and inserts lightweight modules for cross-modal adaptation and interaction. Early approaches mainly explore prompt-based~\cite{lester2021powerscaleparameterefficientprompt} strategies. ProTrack~\cite{yang2022prompting} pioneers this by generating mixed-modality prompts through linear combinations of RGB and X-modal features. The followers~\cite{zhu2023visual,hong2024onetracker,wu2024single} refine this design by treating X-modal inputs as layer-wise prompts, fusing them with RGB features through proposed prompter blocks. Beyond prompt tuning, BAT~\cite{cao2024bi} employs bi-directional adapters~\cite{houlsby2019parameter} for mutual information extraction between RGB and thermal modalities. XTrack~\cite{tan2025xtrack} leverages MoE to separately extract modality-specific information before aggregating them via element-wise addition. Despite the diversity in feature extraction ways, the absence of spatial interaction restricts these methods to performing local fusion with constrained expressiveness. SDSTrack~\cite{hou2024sdstrack} addresses this by reusing frozen attention layers for global information interaction. However, attention mechanisms suffer from limited efficiency due to their quadratic complexity. In response to this limitation, we introduce Hierarchical MoE (HMoE), an efficient global relation modeler for cross-modal fusion. 

\vspace{-4pt}
\subsection{Cross-modal Alignment}
As another fundamental concept in multimodal learning, alignment aims to ensure that different modalities are properly matched and synchronized. CLIP~\cite{radford2021learning} and its follow-up~\cite{li2022blip} achieve this in vision-language tasks through contrastive learning, demonstrating remarkable zero-shot capabilities. In autonomous driving, camera-LiDAR alignment forms the foundation for robust multi-sensor perception~\cite{yin2024fusion, wang2025mambafusionheightfidelitydenseglobal}. Although alignment has proven necessary in these fields, multimodal tracking research focuses primarily on cross-modal fusion strategies~\cite{hu2023transformer,wu2024single,hu2024toward,xiao2025cross,hu2025exploiting,hu2025adaptive,shi2025mamba,shi2025swimvg}. Some works~\cite{zhang2024amnet,li2025cadtrack} explore feature-level spatial alignment but overlook inconsistent intra-modal matching as a critical issue in current two-stream architectures. To this end, we propose Adaptive Mutual Guidance Low-Rank Adaptation (AMG-LoRA) for cross-modal alignment of such responses (attention maps). Extensive experiments demonstrate that effective alignment is promising for breaking the efficiency-performance dilemma.

\section{Method}
In this paper, we propose SEATrack, a simple, efficient, and adaptive multimodal tracker. The overall pipeline of SEATrack is presented in Fig.~\ref{fig:pipeline}, which consists of the frozen foundation tracker and learnable task-specific components for cross-modal interaction.


\subsection{Overall Architecture}
\begin{figure*}[t!]
    \centering
    \includegraphics[width=\linewidth]{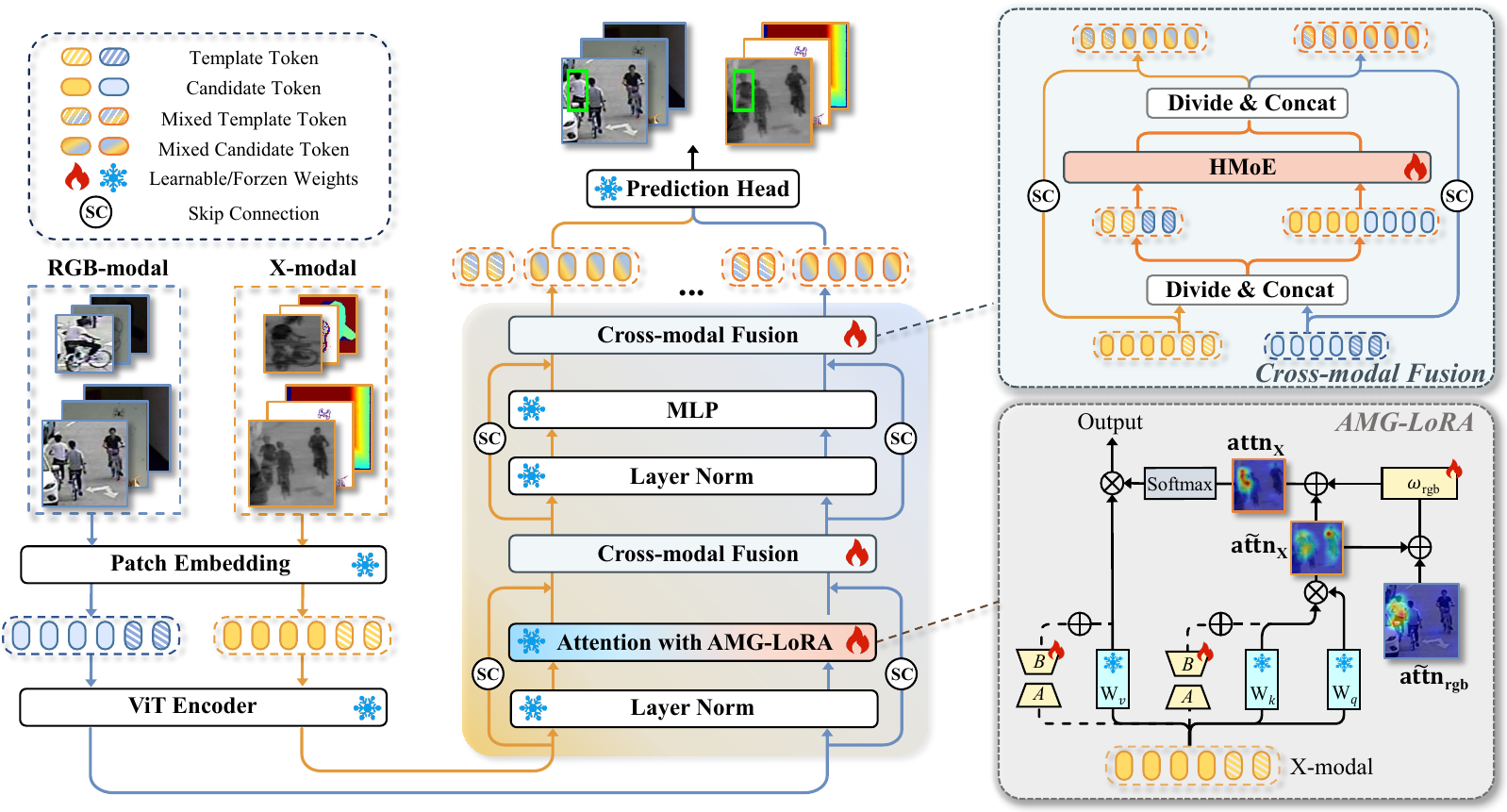}
    \caption{Overall pipeline of SEATrack. Input tokens from each modality are processed by stacked shared ViT encoders for intra-modal target matching and feature extraction. To enable cross-modal interaction, the proposed AMG-LoRA and HMoE are embedded into the ViT encoder every 2 layers to perform attention alignment and information fusion, respectively. AMG-LoRA's mutual guidance mechanism is exemplified through the ``RGB refines X" pathway (Eq.~\ref{eq:r2dte}), with the reverse direction behaving similarly.}
    \label{fig:pipeline}
\end{figure*}

The foundation tracker~\cite{ye2022joint} takes an RGB template-candidate (search region) image pair as input, which is first split into token sequences $\textbf{z}_{\text{rgb}} \in \mathbb{R}^{N_z \times D}$ and $\textbf{c}_{\text{rgb}} \in \mathbb{R}^{N_c \times D}$ via the patch embedding layer. Subsequently, $\textbf{z}_{\text{rgb}}$ and $\textbf{c}_{\text{rgb}}$ are concatenated to form $\textbf{H}_{\text{rgb}} = [\textbf{z}_{\text{rgb}}; \textbf{c}_{\text{rgb}}] \in \mathbb{R}^{(N_z+N_c)\times D}$, which is then fed into stacked ViT encoders for joint feature extraction and matching between the template and the candidate. We copy this pipeline to process the $\text{X}$-modality input, forming the two-stream architecture shown in Fig.~\ref{fig:pipeline} and producing token sequences $\textbf{z}_{\text{X}} \in \mathbb{R}^{N_z \times D}$, $\textbf{c}_{\text{X}} \in \mathbb{R}^{N_c \times D}$, and $\textbf{H}_{\text{X}} = [\textbf{z}_{\text{X}}; \textbf{c}_{\text{X}}] \in \mathbb{R}^{(N_z+N_c)\times D}$ correspondingly. To bridge cross-modal information interaction, we introduce Adaptive Mutual Guidance Low-Rank Adaptation (AMG-LoRA) and Hierarchical MoE (HMoE), embedding them into the ViT encoders every 2 layers. Finally, the extracted candidate features from both modalities are aggregated via element-wise addition and passed to the prediction head for target localization.

\subsection{Cross-modal Attention Alignment}
\label{subsec:amglora}
Template-candidate matching is crucial in tracking, typically accomplished via self-attention to yield the target's position response. With pre-trained attention layers frozen for training efficiency, existing PEFT two-stream methods~\cite{cao2024bi, hou2024sdstrack, tan2024xtrackmultimodaltrainingboosts} suffer from inconsistent matching caused by domain gaps. We argue that, given the multimodal inputs are spatio-temporally aligned, conflicting attention maps derived from the inconsistent matching hinder joint representation learning. To address this, we propose Adaptive Mutual Guidance Low-Rank Adaptation (AMG-LoRA) to facilitate domain adaptation while dynamically refining and aligning matching attention maps across modalities.

To generalize the knowledge learned in the RGB domain to multimodal distributions for improving intra-modal matching, we leverage LoRA~\cite{hu2022lora} to adapt the pre-trained attention weights. As illustrated in Fig.~\ref{fig:pipeline}, LoRA is adopted to update the projection matrices $W_k$ and $W_v$ of the attention layer. During training, it serves as a shared bypass for RGB and X-modal inputs to facilitate joint representation learning. The enhanced matching attention maps can be obtained through:
\begin{gather}
    \tilde{K} = \textbf{H}_*W_k + \textbf{H}_*AB \\
    \tilde{\textbf{attn}}_* = \frac{(\textbf{H}_*W_q)\tilde{K}}{\sqrt{D}}
\end{gather}
where $\textbf{H}_*$ denotes the concatenated tokens from arbitrary modality, $\tilde{\textbf{attn}}_*$ is the corresponding unnormalized (before softmax operation) matching attention map, and $A\in \mathbb{R}^{D\times r},B\in \mathbb{R}^{r\times D}$ are low-rank matrices. During inference, such low-rank matrices can be merged into the original weights for efficient computation.

While domain adaptation of LoRA helps to enhance the matching attention maps, achieving alignment remains challenging because the target salience can differ significantly across modalities under different scene conditions~\cite{cao2024bi, hu2024towards}. Fig.~\ref{fig:vis_amglora} intuitively illustrates how the model's ability to perceive targets in each modality shifts with scene variations. This requires a dynamic and adaptive alignment process to avoid negative transfer, where an unreliable modality may, in turn, misguide the reliable one. Inspired by Classifier-Free Guidance (CFG)~\cite{ho2022classifier}, we tackle this challenge by reformulating such alignment as a multi-branch trade-off problem. Specifically, we propose Adaptive Mutual Guidance (AMG) to dynamically refine and align the attention maps across modalities through a bidirectional learnable linear interpolation, formulated as:
\begin{gather}
    \textbf{attn}_{\text{rgb}} = \tilde{\textbf{attn}}_{\text{rgb}} + w_{\text{X}}(\tilde{\textbf{attn}}_{\text{X}}-\tilde{\textbf{attn}}_{\text{rgb}}) \label{eq:dte2r}  \\
    \textbf{attn}_{\text{X}} = \tilde{\textbf{attn}}_{\text{X}} + w_{\text{rgb}}(\tilde{\textbf{attn}}_{\text{rgb}}-\tilde{\textbf{attn}}_{\text{X}}) \label{eq:r2dte}
\end{gather}
where $w_\text{X}$ and $w_\text{rgb}$ are learnable scaling factors to adapt to the dynamic challenge we discussed. During training, they are initialized to 1 for cross-guidance.

Despite simplicity, AMG-LoRA offers valuable insight that cross-modal attention alignment is promising for breaking the performance-efficiency dilemma. As presented in Tab.~\ref{tab:component_ablation}, it delivers substantial improvements (18.3\%, 7.2\%, and 6.1\% PR gains on LasHeR, DepthTrack, and VisEvent, respectively) with only 0.14M parameters. Moreover, comprehensive quantitative and qualitative evaluations confirm that AMG-LoRA is a well-targeted and effective design for multimodal tracking.

\subsection{HMoE for Cross-modal Fusion}
\begin{figure}[t!]
    \centering
    \includegraphics[width=\linewidth]{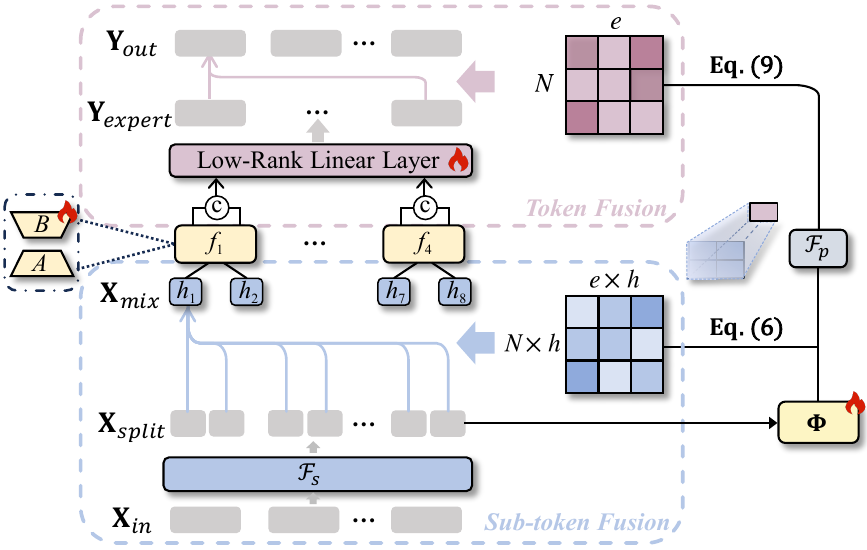}
    \caption{Architecture details of HMoE configured with $e=4$ experts, each containing $h=2$ heads.
    }
    \label{fig:MMoE}
\end{figure}

As discussed before, designing a cross-modal fusion strategy that balances expressiveness and efficiency is a long-standing challenge in multimodal tracking. In response to this, we propose Hierarchical MoE (HMoE), with the details presented in Fig.~\ref{fig:MMoE}. Distinct from existing MoE explorations in tracking~\cite{chen2024emoetrackerenvironmentalmoebasedtransformer, cai2025spmtrack, tan2025xtrack} that focus on expert-level ensemble learning for feature enhancement, HMoE is designed as an efficient global relation modeler. It allocates fine-grained sub-token mixtures to different expert heads and aggregates the resulting expert tokens into the final output through a hierarchical soft routing mechanism. Architecturally, HMoE is inserted after the Attention and FFN sub-layers, where it processes on either the template or candidate token sequences to enable cross-modal fusion, as illustrated in Fig.~\ref{fig:pipeline}.


Let $\textbf{X}_{in}\in \mathbb{R}^{N\times D}$ denote the input token sequence consisting of $N$ tokens, each with $D$ dimensions. As mentioned before, $\textbf{X}_{in}$ can be either $[\textbf{z}_{\text{rgb}}; \textbf{z}_{\text{X}}]$ or $[\textbf{c}_{\text{rgb}}; \textbf{c}_{\text{X}}]$. For parameter efficiency, HMoE uses low-rank linear layer as expert function and the $e$ experts are defined as $\{f_i: \mathbb{R}^{\frac{D}{h}} \rightarrow \mathbb{R}^{r} \rightarrow \mathbb{R}^{\frac{D}{h}}\}_{i=1}^e$, with $r\ll \frac{D}{h}$. In HMoE, each expert has $h$ heads to process, each associated with a learnable $\frac{D}{h}$-dimensional vector that measures the affinity between the input and the head, forming the learnable gating matrix $\boldsymbol{\Phi} \in \mathbb{R}^{\frac{D}{h} \times (e \cdot h)}$.

The input $\textbf{X}_{in}$ is first transformed by a low-rank linear layer. Each token of the output is then split into $h$ sub-tokens, forming $\textbf{X}_{split} \in \mathbb{R}^{(N\cdot h) \times (\frac{D}{h})}$, as formulated by:
\begin{gather}
    \textbf{X}_{split}=\mathcal{F}_s (\textbf{X}_{in})
\end{gather}
where $\mathcal{F}_s$ is channel split operation. 


Guided by the gating matrix $\boldsymbol{\Phi}$, sub-tokens from $\textbf{X}_{split}$ are mixed to generate head-level inputs $\textbf{X}_{mix} \in \mathbb{R}^{(e\cdot h) \times \frac{D}{h}}$ for each expert. Subsequently, outputs from the same expert are merged into expert-level tokens, followed by another low-rank linear layer to yield $\textbf{Y}_{expert} \in \mathbb{R}^{e\times D}$. This sub-token fusion process can be formulated as:
\begin{gather}
    \textbf{X}_{mix} = \mathrm{softmax}(\textbf{X}_{split} \boldsymbol{\Phi},\ \text{dim=0})^{\textbf{T}} \textbf{X}_{split} \label{eq:6}\\[0.5ex]
    \textbf{Y}^{i,j}_{head} = f_i\left(\textbf{X}^{i,j}_{mix}\right),i \in \{1,\ldots,e\},\ j \in \{1,\ldots,h\} \label{eq:7}\\[0.5ex]
    \textbf{Y}_{expert} = \mathcal{F}_m(\textbf{Y}_{head})
\end{gather}
where $\textbf{Y}_{head}\in \mathbb{R}^{(e\cdot h)\times \frac{D}{h}}$ is the head-level activations by experts and $\mathcal{F}_m$ denotes the channel concatenate operation.

Token-level fusion, analogous to Eq.~\ref{eq:6} but requires a token-to-expert affinity matrix for combining expert tokens into final outputs. Note that each row of $\textbf{X}_{split} \boldsymbol{\Phi}\in \mathbb{R}^{(N\cdot h)\times (e\cdot h)}$ represents the affinities of a sub-token to all heads, and each column corresponds to the affinities of all sub-tokens to a single head, we can reconstruct the token-to-expert affinity matrix ${\textbf{A}}\in \mathbb{R}^{N\times e}$ by aggregating non-overlapping $h\times h$ patches. Therefore, the final output $\textbf{Y}_{out}\in \mathbb{R}^{N\times D}$ can be obtained by:
\begin{gather}
    \textbf{A} = \mathrm{softmax}\left(\mathcal{F}_{p}(\textbf{X}_{split}\boldsymbol{\Phi}),\ \text{dim=1}\right) \label{eq:9}\\
    \textbf{Y}_{out} = \textbf{A}\textbf{Y}_{expert}\label{eq:10}
\end{gather}
where $\mathcal{F}_{p}: \mathbb{R}^{(N\cdot h)\times (e\cdot h)}\rightarrow \mathbb{R}^{N\times e}$ is patchify operation. Through hierarchical fusion progressing from the sub-token level (Eq.~\ref{eq:6}) to the token level (Eq.~\ref{eq:9}), target cues across modalities are effectively enriched. 

\subsection{Prediction Head and Loss}
We inherit the head from the foundational tracker to locate the target within the candidate. The summation of $\textbf{c}_{\text{rgb}}$ and $\textbf{c}_{\text{X}}$ is fed into the head. Then, three stacked Conv-BN-ReLU blocks are followed to generate the target center score map, center coordinate offset, and bounding box size, respectively. We adopt the same objective loss as the foundation tracker, given by:
\begin{equation}
L=L_{focal}+\lambda_{iou}L_{iou}+\lambda_{L1}L_1
\end{equation}
where $\lambda_{iou}=2$ and $\lambda_{L1}=5$ are the regularization factors.

\section{Experiment}
\label{Sec:Exp}
\begin{table*}[t!]
    \centering
    \caption{Main performance comparison across RGB-T, RGB-D, and RGB-E datasets. SEATrack delivers strong overall results on five benchmarks, with outstanding parameter efficiency. L.P. denotes Learnable Parameters. \textsuperscript{*} indicates reproduced results.
    }
    \label{tab:main}
    \resizebox{1\linewidth}{!}{
    \begin{tabular}{c|c|c|ccc|cc|ccc|ccc|cc|c|c}
        \toprule
        \multicolumn{1}{c|}{} &
        \multicolumn{1}{c|}{} &
        \multicolumn{1}{c|}{} &
        \multicolumn{3}{c|}{\textbf{LasHeR}} & 
        \multicolumn{2}{c|}{\textbf{RGBT234}} & 
        \multicolumn{3}{c|}{\textbf{DepthTrack}} & 
        \multicolumn{3}{c|}{\textbf{VOT-RGBD2022}} & 
        \multicolumn{2}{c|}{\textbf{VisEvent}} & 
        \multicolumn{1}{c|}{} &
        \multicolumn{1}{c}{}\\
        \cline{4-16}
            \multirow{-2}{*}{\textbf{Type}} &
            \multirow{-2}{*}{\textbf{Method}} & 
            \multirow{-2}{*}{\textbf{Source}} & 
            \multirow{1.3}*{PR\textuparrow} & 
            \multirow{1.3}*{NPR\textuparrow} & 
            \multirow{1.3}*{SR\textuparrow} & 
            \multirow{1.3}*{MPR\textuparrow} & 
            \multirow{1.3}*{MSR\textuparrow} & 
            \multirow{1.3}*{PR\textuparrow} & 
            \multirow{1.3}*{RE\textuparrow} & 
            \multirow{1.3}*{F-score\textuparrow} & 
            \multirow{1.3}*{EAO\textuparrow} & 
            \multirow{1.3}*{Acc.\textuparrow} & 
            \multirow{1.3}*{Rob.\textuparrow} & 
            \multirow{1.3}*{PR\textuparrow} & 
            \multirow{1.3}*{SR\textuparrow} &
            \multirow{-2}{*}{\textbf{L.P.}\text{\textdownarrow}} & 
            \multirow{-2}{*}{\textbf{FPS}\textuparrow}\\
        \midrule
        \multirow{7}{*}{\makecell{\textbf{FFT}}}
        & TBSI~\cite{hui2023bridging} & CVPR'23 & 69.2 & 65.7 & 55.6 & 87.1 & 63.7 & - & - & - & - & - & - & - & - & 203M & 36.2\\
        & IIMF~\cite{chen2024simplifying} & MM'24 & 72.4 & 68.4 & 58.1 & 86.8 & 64.4 & - & - & - & - & - & - & - & - & 182M & -\\
        & CAFormer~\cite{xiao2025cross} & AAAI'25 & 70.0 & 66.1 & 55.6 & 88.3 & 66.4 & - & - & - & - & - & - & - & - & - & 83.6\\
        & DeT~\cite{yan2021depthtrack} & CVPR'21 & - & - & - & - & - & 56.0 & 50.6 & 53.2 & 65.7 & 76.0 & 84.5 & - & - & - & -\\
        & TABBTrack~\cite{ying2025temporal} & PR'25 & - & - & - & - & - & 62.2 & 61.5 & 61.8 & 72.2 & 82.1 & 87.4 & - & - & - & 27\\
        & PHPTrack~\cite{wang2025prior} & TCSVT'25 & - & - & - & - & - & - & - & - & - & - & - & 77.8 & 61.1 & 93.7M & 21.5\\
        & CMDTrack~\cite{zhang2025cross} & TPAMI'25 & 68.8 & - & 56.6 & 85.9 & 61.8 & 59.1 & 60.7 & 59.8 & - & - & - & 75.8 & 61.3 & 31M & 67\\
        & MamTrack~\cite{sun2025exploring} & CVPR'25 & 67.4 & - & 54.2 & 84.4 & 62.4 & - & - & - & - & - & - & 79.2 & 61.6 & - & 18.1\\
        & SMSTrack~\cite{chan2025smstracker} & ICCV'25 & 70.3 & - & 56.0 & 86.9 & 64.5 & 64.1 & 63.1 & 63.6 & 74.8 & 82.2 & 89.7 & 76.3 & 60.4 & 98.2M\textsuperscript{*} & 24.1\\
        \midrule
        \multirow{14}{*}{\makecell{\textbf{PEFT}}}
        & ProTrack~\cite{yang2022prompting} & MM'22 & 53.8 & 49.8 & 42.0 & 79.5 & 59.9 & 58.3 & 57.3 & 57.8 & 65.1 & 80.1 & 80.2 & 63.2 & 47.1 & - & -\\
        & ViPT~\cite{zhu2023visual} & CVPR'23 & 65.1 & 61.6 & 52.5 & 83.5 & 61.7 & 59.2 & 59.6 & 59.4 & 72.1 & 81.5 & 87.1 & 75.8 & 59.2 & 0.8M & -\\
        & VADT~\cite{zhang2024visual} & ICASSP'24 & - & - & - & - & - & 60.3 & 60.6 & 61.0 & 72.1 & 81.6 & 87.3 & - & - & - & -\\
        & TATrack~\cite{wang2024temporal} & AAAI'24 & 70.2 & 66.7 & 56.1 & 87.2 & 64.4 & - & - & - & - & - & - & - & - & - & 26.1\\
        & BAT~\cite{cao2024bi} & AAAI'24 & 70.2 & 66.4 & 56.3 & 86.8 & 64.1 & - & - & - & - & - & - & - & - & 0.3M & -\\
        & eMoET~\cite{chen2024emoetrackerenvironmentalmoebasedtransformer} & RAL'24 & - & - & - & - & - & - & - & - & - & - & - & 76.4 & 61.3 & 8.4M & -\\
        & UnTrack~\cite{wu2024single} & CVPR'24 & 64.6 & 60.0 & 51.3 & 84.2 & 62.5 & 61.1 & 60.8 & 61.0 & 72.1 & 82.0 & 86.9 & 75.5 & 58.9 & 6.6M & 25.6\\
        & OneTracker~\cite{hong2024onetracker} & CVPR'24 & 67.2 & - & 53.8 & 85.7 & 64.2 & 60.7 & 60.4 & 60.9 & 72.7 & 81.9 & 87.2 & 76.7 & 60.8 & 2.8M & -\\
        & SDSTrack~\cite{hou2024sdstrack} & CVPR'24 & 66.5 & 62.6 & 53.1 & 84.8 & 62.5 & 61.9 & 60.9 & 61.4 & 72.8 & 81.2 & 88.3 & 76.7 & 59.7 & 14.8M & 20.8\\
        & SNNPTrack~\cite{sun2025exploring} & ICASSP'25 & - & - & - & - & - & - & - & - & - & - & - & 76.9 & 59.8 & 2.1M & -\\
        & M\textsuperscript{3}Track~\cite{tang2025m} & SPL'25 & 65.8 & - & 52.5 & 84.5 & 63.0 & 56.4 & 58.5 & 57.4 & - & - & - & 76.7 & 59.6 & 1.7M & 31.1\\
        & XTrack~\cite{tan2025xtrack} & ICCV'25 & 69.1 & 65.5 & 55.7 & 87.4 & 64.9 & 61.8 & 62.0 & 61.5 & 74.0 & 82.1 & 88.8 & 77.5 & 60.9 & 5.4M\textsuperscript{*} & 10.3\textsuperscript{*}\\
        \cmidrule(){2-18}
        & \cellcolor{cyan!15}SEATrack & \cellcolor{cyan!15}Ours & \cellcolor{cyan!15}\textbf{71.6} & \cellcolor{cyan!15}\textbf{67.5} & \cellcolor{cyan!15}\textbf{57.3} & \cellcolor{cyan!15}\textbf{87.8} & \cellcolor{cyan!15}\textbf{63.9} & \cellcolor{cyan!15}\textbf{62.9} & \cellcolor{cyan!15}\textbf{63.5} & \cellcolor{cyan!15}\textbf{63.2} & \cellcolor{cyan!15}\textbf{73.6} & \cellcolor{cyan!15}\textbf{82.1} & \cellcolor{cyan!15}\textbf{88.4} & \cellcolor{cyan!15}\textbf{77.1} & \cellcolor{cyan!15}\textbf{60.3} & \cellcolor{cyan!15}\textbf{0.6M} & \cellcolor{cyan!15}\textbf{63.5}\\
        \bottomrule
    \end{tabular}
}
    \vspace{-10pt}
\end{table*}

\subsection{Implementation Details}
Follow common practice~\cite{zhu2023visual, hong2024onetracker, wu2024single, hou2024sdstrack,chan2025smstracker,tan2025xtrack}, SEATrack builds upon the ViT-Base version of OSTrack~\cite{ye2022joint} as the foundation tracker and is fine-tuned on two 4090 GPUs with a global batch size of 64 using PyTorch. SEATrack takes a $128\times128$ template and a $256\times256$ search region as input, with training performed on LasHeR~\cite{li2021lasher}, DepthTrack~\cite{yan2021depthtrack}, and VisEvent~\cite{wang2023visevent} for RGB-T, RGB-D, and RGB-E tracking, respectively. We employ the AdamW optimizer with a weight decay of $10^{-4}$. The learning rate is set to $4 \times 10^{-4}$ for RGB-T and RGB-D, and $6 \times 10^{-5}$ for RGB-E. The training is conducted for 60, 25, and 45 epochs on RGB-T, RGB-D, and RGB-E tracking, respectively. AMG-LoRA and HMoE are inserted into the ViT encoder every 2 layers with Xavier initialization, and only their parameters are updated during training while the foundation tracker remains frozen. Efficiency-related metrics are tested on RTX 4090, SEATrack runs 63.5 FPS with about 1GB of memory usage.

\subsection{Comparison with State-of-the-arts}

\paragraph{LasHeR.}LasHeR~\cite{li2021lasher} is the most popular high-diversity benchmark for RGB-T tracking that contains 245 test sequences and 975 train sequences.The precision (PR), Normalized Precision Rate (NPR), and success (SR) metrics are used to evaluate tracking performance. As shown in Tab.~\ref{tab:main}, SEATrack outperforms all existing PEFT-based trackers and remains highly competitive compared to the FFT-based method IIMF~\cite{chen2024simplifying}, with only about 0.3\% of its parameters.
\vspace{-12pt}

\paragraph{RGBT234.}RGBT234~\cite{li2019rgb} is a large-scale RGB-T tracking benchmark, which contains 234 videos. To reduce the influence of annotation-induced alignment errors~\cite{li2019rgb}, Maximum Precision Rate (MPR) and Maximum Success Rate (MSR) are adopted for evaluation rather than the conventional PR and SR metrics. Tab.~\ref{tab:main} demonstrates that SEATrack outperforms all PEFT-based methods in MPR while achieving competitive results against FFT-based approaches. Despite a 1\% MSR gap relative to XTrack\cite{tan2025xtrack}, SEATrack's superior parameter efficiency (5.4M \textit{\textbf{vs.}} 0.6M) renders it substantially more resource-friendly.
\vspace{-12pt}

\paragraph{DepthTrack.}DepthTrack~\cite{yan2021depthtrack} is a large-scale long-term RGB-D tracking benchmark with 150 training and 50 testing video sequences. The metrics precision (PR) and recall (RE) are used to measure the accuracy and robustness of target localization. F-score, the primary measurement, is calculated by $\text{F}=\frac{2\text{PR}\cdot \text{RE}}{\text{PR}+\text{RE}}$. As shown in Tab.~\ref{tab:main}, SEATrack consistently outperforms all PEFT-based trackers.
\vspace{-12pt}

\paragraph{VOT-RGBD2022.}VOT-RGBD2022~\cite{kristan2022tenth} is the most recent benchmark for RGB-D tracking, which contains 127 RGB-D video sequences. Expected Average Overlap (EAO), Accuracy, and Robustness are used to evaluate performance. As shown in Tab.~\ref{tab:main}, SEATrack surpasses SDSTrack~\cite{hou2024sdstrack} by 0.8\% in EAO, 0.9\% in Accuracy, and 0.1\% in Robustness, with a clear advantage in efficiency.

\vspace{-5pt}
\paragraph{VisEvent.}As the most representative benchmark for RGB-E tracking, VisEvent~\cite{wang2023visevent} contains 500 sequences for training and 320 sequences for testing, both collected from real-world scenarios. Similar to the LasHeR~\cite{li2021lasher}, precision (PR) and success (SR) are used to evaluate tracking performance. As shown in Tab.~\ref{tab:main}, SEATrack, as a unified approach, achieves performance comparable to the task-specific method ~\cite{chen2024emoetrackerenvironmentalmoebasedtransformer}. More importantly, it is also competitive with FFT-based SOTA~\cite{chan2025smstracker}.

\begin{table}[t]
    \centering
    \caption{Component-wise Ablation Studies.}
    \label{tab:component_ablation}
    \resizebox{\linewidth}{!}{
    \begin{tabular}{cc|cc|ccc|cc|c}
        \toprule
        \multicolumn{1}{c}{\multirow{2}{*}{\textbf{AMG-LoRA}}} &
        \multicolumn{1}{c|}{\multirow{2}{*}{\textbf{HMoE}}} & 
        \multicolumn{2}{c|}{\textbf{LasHeR}} & 
        \multicolumn{3}{c|}{\textbf{DepthTrack}} & 
        \multicolumn{2}{c|}{\textbf{VisEvent}} & 
        \multicolumn{1}{c}{\multirow{2}{*}{\textbf{Param.}}} \\
        \cline{3-9}
            & & 
            \multirow{1.3}*{PR} & 
            \multirow{1.3}*{SR} & 
            \multirow{1.3}*{PR} & 
            \multirow{1.3}*{RE} & 
            \multirow{1.3}*{F-score} & 
            \multirow{1.3}*{PR} & 
            \multirow{1.3}*{SR} & \\
        \midrule
        - & - & 51.5 & 41.2 & 53.6 & 52.2 & 52.9 & 69.5 & 53.4 & 0 \\
        \midrule
        \checkmark & - & 69.8 & 55.7 & 58.1 & 57.2 & 57.6 & 75.6 & 59.1 & 0.14M \\
        - & \checkmark & 67.4 & 54.2 & 60.9 & 61.3 & 61.1 & 76.5 & 59.9 & 0.46M \\
        \midrule
        \rowcolor{cyan!15} \checkmark & \checkmark & \textbf{71.6} & \textbf{57.3} & \textbf{62.9} & \textbf{63.5} & \textbf{63.2} & \textbf{77.1} & \textbf{60.3} & \textbf{0.6M} \\
        \bottomrule
    \end{tabular}
    }
    \vspace{-10pt}
\end{table}

\subsection{Ablation and Discussion}
\paragraph{Component Analysis.}As shown in Tab.~\ref{tab:component_ablation}, both components demonstrate significant improvements over the baseline across three tasks, with their combination achieving optimal performance. The baseline is a two-stream variant of the frozen foundation tracker~\cite{ye2022joint} without cross-modal interaction in the ViT encoders, and the two search regions are simply fused by element-wise addition before the prediction head. Notably, both AMG-LoRA and HMoE individually outperform well-established methods~\cite{zhu2023visual,wu2024single,hong2024onetracker,hou2024sdstrack} on LasHeR, while maintaining competitive on DepthTrack and VisEvent. The results of AMG-LoRA, in particular, reveal the potential of cross-modal attention alignment for breaking the performance–efficiency dilemma.

\vspace{-12pt}
\paragraph{Initialization of the Scaling Factor.} The learnable scaling factors $w_{\text{X}}$ (Eq.~\ref{eq:dte2r}) and $w_{\text{rgb}}$ (Eq.~\ref{eq:r2dte}) determine the bidirectional guidance strength of AMG-LoRA. In Tab.~\ref{tab:amg_init}, we ablate three representative initialization strategies and analyze the impact on performance. As a stable starting point, 0-initialization makes the early stage of training degrade into a no-guidance behavior, but the performance appears to be unsatisfactory. Another intuitive choice is to treat both modalities equally, \textit{i.e.}, 0.5-initialization, but the performance under this prior is still not optimal. Benefiting from the cross-guidance induced by 1-initialization, the RGB branch’s matching responses provide a strong prior for X-modal representation learning, yielding the best performance across all three tasks.
\vspace{-12pt}

\begin{figure}[t!]
    \centering
    \includegraphics[width=\linewidth]{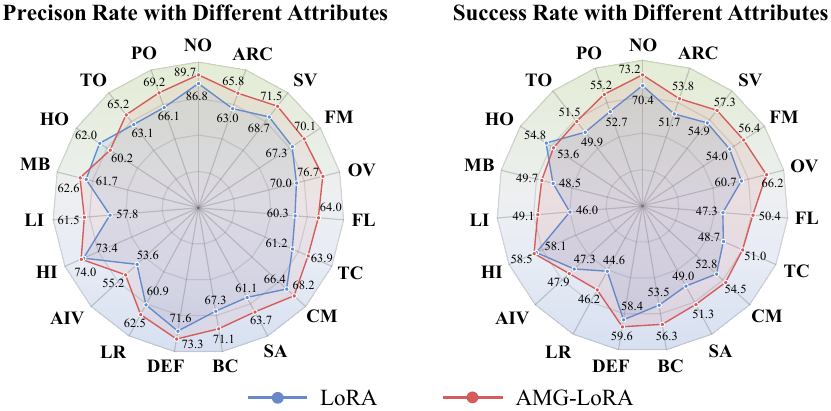}
    \caption{\textbf{LoRA} \textit{v.s.} \textbf{AMG-LoRA} across 19 challenging attributes on LasHeR~\cite{li2021lasher}.
    }
    \label{fig:amglora_vs_lora}
    \vspace{-16pt}
\end{figure}

\begin{figure}[t!]
    \centering
    \includegraphics[width=\linewidth]{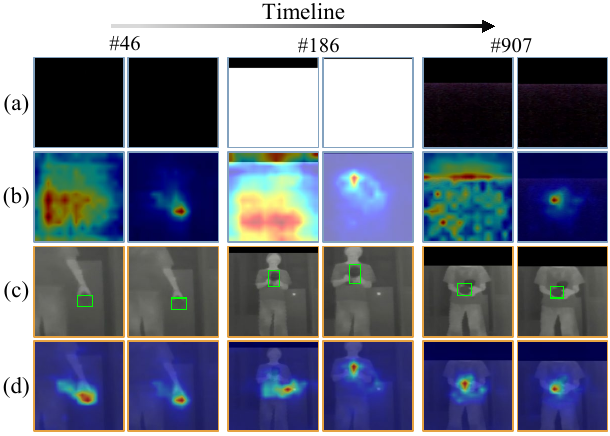}
    \vspace{-15pt}
    \caption{\textbf{LoRA} (left) \textit{vs.} \textbf{AMG-LoRA} (right) under a modality-missing scenario. (a) Missing RGB frame. (b) RGB attention map. (c) Thermal frame. (d) Thermal attention map. 
    }
    \label{fig:modality-missing_vis}
    \vspace{-16pt}
\end{figure}

\paragraph{Deeper Analysis of AMG-LoRA.}To investigate how AMG-LoRA performs in complex conditions, we adopt LoRA~\cite{hu2022lora} as the baseline and evaluate both methods across 19 challenging attributes from the LasHeR~\cite{li2021lasher}. As shown in Fig.~\ref{fig:amglora_vs_lora}, AMG-LoRA achieves non-trivial improvements over LoRA across nearly all attributes. For general scenarios such as Similar Appearance (SA), Background Clutter (BC), and Fast Motion (FM), AMG-LoRA achieves PR–SR gains of 2.6–2.3\%, 3.8–2.8\%, and 2.8–2.4\%, respectively, highlighting its contribution to real-world robustness. More notably, for Out-of-View (OV) and Frame-Loss (FL) scenarios where certain modalities are unavailable and thus contradict our design assumption, AMG-LoRA surprisingly outperforms LoRA in both PR (by 6.7 and 3.7 points) and SR (by 5.5 and 3.1 points). Combined with the results in Fig.~\ref{fig:modality-missing_vis}, which serves as an example of FL, we believe such improvement can be attributed to the benefits of alignment.
\vspace{-24pt}

\begin{table}[t]
    \centering
    \caption{Initialization Choices of AMG-LoRA's Scaling Factor.}
    \label{tab:amg_init}
    \vspace{-5pt}
    \resizebox{!}{31pt}{
    \begin{tabular}{c|cc|ccc|cc}
        \toprule
        \multicolumn{1}{c|}{\multirow{2}{*}{$w_{\text{X}}$ \textbf{\&} $w_{\text{rgb}}$}} &
        \multicolumn{2}{c|}{\textbf{LasHeR}} & 
        \multicolumn{3}{c|}{\textbf{DepthTrack}} & 
        \multicolumn{2}{c}{\textbf{VisEvent}} \\
        \cline{2-8}
            & 
            \multirow{1.3}*{PR} & 
            \multirow{1.3}*{SR} & 
            \multirow{1.3}*{PR} & 
            \multirow{1.3}*{RE} & 
            \multirow{1.3}*{F-score} & 
            \multirow{1.3}*{PR} & 
            \multirow{1.3}*{SR}\\
        \midrule
        0 & 70.4 & 56.2 & 57.5 & 57.6 & 57.5 & 75.9 & 59.6 \\
        \midrule
        0.5 & 69.7 & 55.7 & 58.4 & 59.5 & 58.9 & 76.8 & 60.2 \\
        \midrule
        \rowcolor{cyan!15} \textbf{1} & \textbf{71.6} & \textbf{57.3} & \textbf{62.9} & \textbf{63.5} & \textbf{63.2} & \textbf{77.1} & \textbf{60.3} \\
        \bottomrule
    \end{tabular}
    }
\end{table}

\paragraph{Number of Heads in HMoE.}The number of heads per expert determines the sub-token dimensionality. As shown in Tab.~\ref{tab:head_number_HMoE}, using 2 heads per expert achieves optimal performance. With only 1 head per expert, HMoE's hierarchical fusion degrades to token-level fusion~\cite{puigcerver2024sparsesoftmixturesexperts}, where the large latent space with noise may lead to poor performance. Continuously increasing the number of heads is also harmful, as the over-splitting disrupts the original semantic information.
\vspace{-12pt}

\begin{table}[t]
    \centering
    \caption{Different Numbers of Heads per Expert in HMoE.}
    \label{tab:head_number_HMoE}
    \vspace{-5pt}
    \resizebox{\linewidth}{!}{
    \begin{tabular}{c|c|c|cc|ccc|cc}
        \toprule
        \multicolumn{1}{c|}{\multirow{2}{*}{\textbf{Heads}}} &
        \multicolumn{1}{c|}{\multirow{2}{*}{\textbf{Dim.}}} &
        \multicolumn{1}{c|}{\multirow{2}{*}{\textbf{Param.}}} &
        \multicolumn{2}{c|}{\textbf{LasHeR}} & 
        \multicolumn{3}{c|}{\textbf{DepthTrack}} & 
        \multicolumn{2}{c}{\textbf{VisEvent}}\\
        \cline{4-10}
             & & & 
             \multirow{1.3}*{PR} & 
             \multirow{1.3}*{SR} & 
             \multirow{1.3}*{PR} & 
             \multirow{1.3}*{RE} & 
             \multirow{1.3}*{F-score} & 
             \multirow{1.3}*{PR} & 
             \multirow{1.3}*{SR}\\
        \midrule
        1 & 768 & 0.9M & 69.7 & 55.8 & 58.1 & 59.7 & 58.9 & 76.7 & 60.2 \\
        \rowcolor{cyan!15}\textbf{2} & \textbf{384} & \textbf{0.6M} & \textbf{71.6} & \textbf{57.3} & \textbf{62.9} & \textbf{63.5} & \textbf{63.2} & \textbf{77.1} & \textbf{60.3} \\
        4 & 192 & 0.5M & 70.6 & 56.4 & 60.7 & 58.5 & 59.6 & 76.6 & 59.9\\
        8 & 96 & 0.4M & 70.6 & 56.5 & 61.0 & 61.7. & 61.3 & 76.3 & 59.7\\
        \bottomrule
    \end{tabular}
    }
\end{table}

\begin{table}[t!]
    \centering
    \caption{Different Numbers of Experts in HMoE.}
    \label{tab:expert_number_HMoE}
    \vspace{-5pt}
    \resizebox{\linewidth}{!}{
    \begin{tabular}{c|c|c|cc|ccc|cc}
        \toprule
        \multicolumn{1}{c|}{\multirow{2}{*}{\textbf{Attn.}}} &
        \multicolumn{1}{c|}{\multirow{2}{*}{\textbf{FFN.}}} &
        \multicolumn{1}{c|}{\multirow{2}{*}{\textbf{Param.}}} &
        \multicolumn{2}{c|}{\textbf{LasHeR}} & 
        \multicolumn{3}{c|}{\textbf{DepthTrack}} & 
        \multicolumn{2}{c}{\textbf{VisEvent}}\\
        \cline{4-10}
             & & & 
             \multirow{1.3}*{PR} & 
             \multirow{1.3}*{SR} & 
             \multirow{1.3}*{PR} & 
             \multirow{1.3}*{RE} & 
             \multirow{1.3}*{F-score} & 
             \multirow{1.3}*{PR} & 
             \multirow{1.3}*{SR}\\
        \midrule
        4 & 4 & 0.5M & 70.3 & 56.0 & 60.9 & 61.6 & 61.2 & 76.8 & 60.2 \\
        8 & 4 & 0.6M & 70.7 & 56.6 & 59.7 & 60.0 & 59.8 & 76.6 & 59.9 \\
        \rowcolor{cyan!15}\textbf{4} & \textbf{8} & \textbf{0.6M} & \textbf{71.6} & \textbf{57.3} & \textbf{62.9} & \textbf{63.5} & \textbf{63.2} & \textbf{77.1} & \textbf{60.3}\\
        8 & 16 & 0.9M & 70.9 & 56.7 & 59.3 & 59.0 & 59.1 & 76.1 & 59.7 \\
        \bottomrule
    \end{tabular}
    }
\end{table}

\begin{figure*}[t!]
    \centering
    \includegraphics[width=\linewidth]{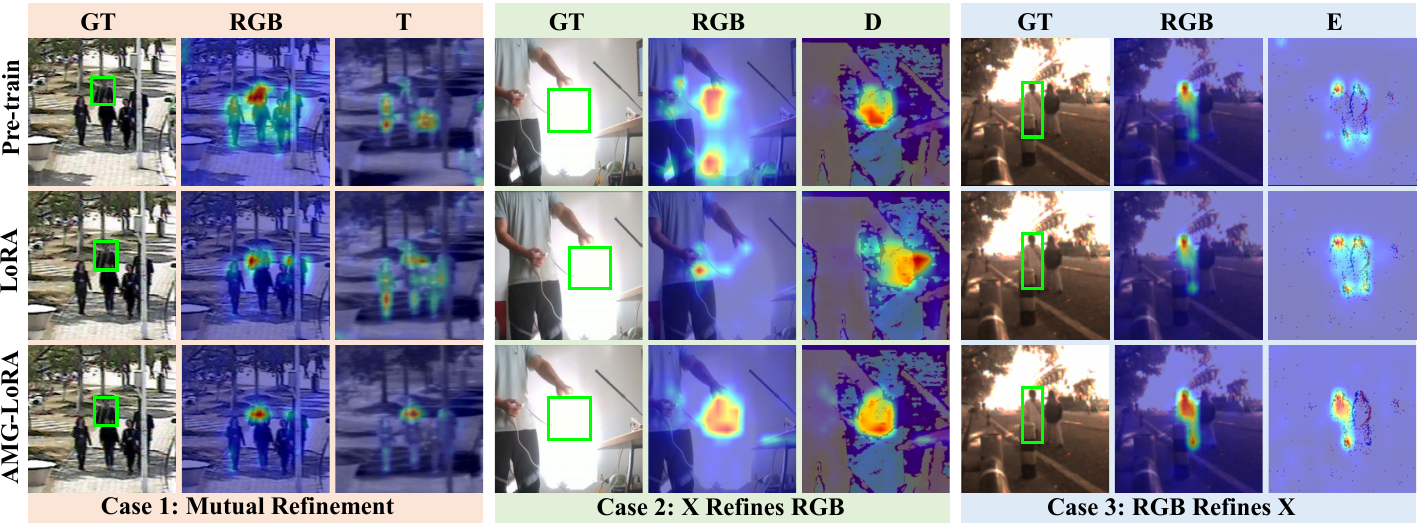}
    \caption{Comprehensive visualization of AMG-LoRA's adaptability. The results of ``Pre-train" row are directly inferred from the frozen foundation tracker~\cite{ye2022joint}.}
    \label{fig:vis_amglora}
    \vspace{-6pt}
\end{figure*}

\paragraph{Number of Experts in HMoE.}The number of experts is another critical hyperparameter in HMoE, as it directly determines both the model’s capacity and parameter size. Tab.~\ref{tab:expert_number_HMoE} demonstrates that prioritizing expert budget to FFN layers over attention layers yields optimal performance across all three tasks, consistent with observations in~\cite{he2021towards}. Furthermore, excessively increasing the number of experts under limited data conditions leads to overfitting and subsequent performance degradation.
\vspace{-12pt}

\paragraph{Comparison with Different Fusion Strategies.}To validate HMoE's design philosophy of balancing expressiveness and efficiency, we compare it with two representative methods: cross-attention (expressiveness-oriented through global fusion) and MCP~\cite{zhu2023visual} (efficiency-oriented through local fusion). Both follow HMoE's template-to-template and candidate-to-candidate fusion strategy and identical insertion positions, while we extend them to support bidirectional fusion. As shown in Tab.~\ref{tab:fusion_compare}, HMoE achieves very comparable performance over the cross-attention while achieving about 35\% speed-up in FPS. The extended bidirectional MCP surprisingly leads to LasHeR, yet it sacrifices efficiency due to the multi-turn operations required. In contrast, HMoE enjoys high efficiency owing to its inherent global receptive field. Moreover, AMG-LoRA boosts all three fusion strategies across most metrics while introducing negligible additional latency, demonstrating its good generalization. In particular, combining AMG-LoRA with HMoE yields both optimal performance and efficiency, providing compelling justification for our design choice.
\vspace{-12pt}

\begin{figure}[t!]
    \centering
    \includegraphics[width=\linewidth]{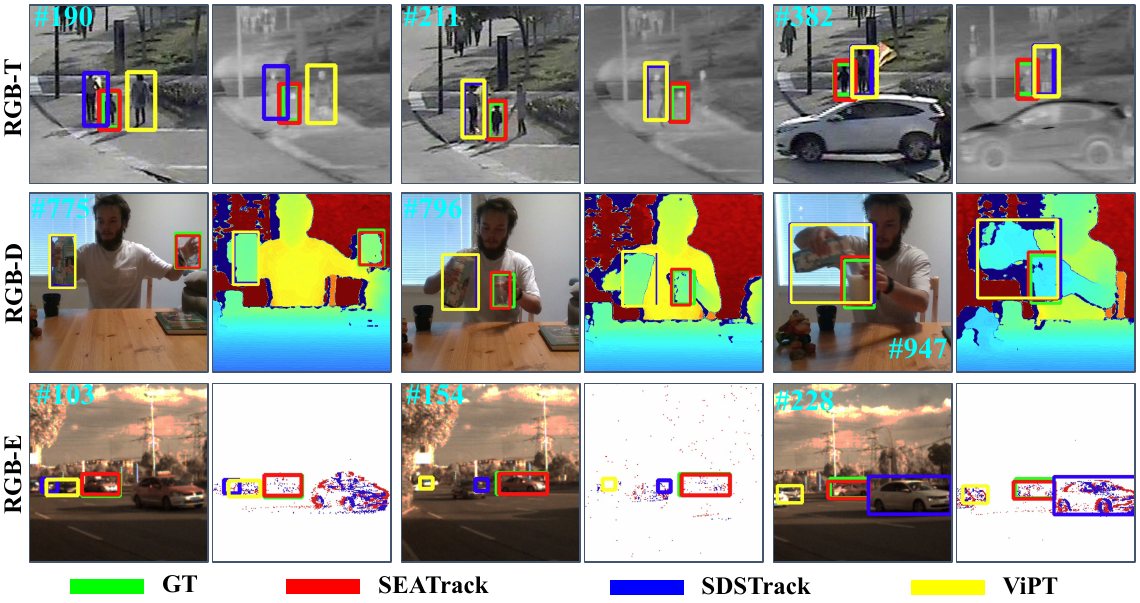}
    \vspace{-16pt}
    \caption{Prediction-level comparison of SEATrack with two well-established PEFT-based multimodal trackers.}
    \label{fig:vis_bbox}
    \vspace{-15pt}
\end{figure}

\begin{table}[t!]
    \centering
    \caption{Comparison of Different Fusion Strategies.}
    \label{tab:fusion_compare}
    \vspace{-5pt}
    \resizebox{\linewidth}{!}{
    \begin{tabular}{c|cc|ccc|cc|c|c}
        \toprule
        \multicolumn{1}{c|}{\multirow{2}{*}{\textbf{Variants}}} & \multicolumn{2}{c|}{\textbf{LasHeR}} & 
        \multicolumn{3}{c|}{\textbf{DepthTrack}} & 
        \multicolumn{2}{c|}{\textbf{VisEvent}} & 
        \multicolumn{1}{c|}{\multirow{2}{*}{\textbf{MACs (10\textsuperscript{9})}}} & 
        \multicolumn{1}{c}{\multirow{2}{*}{\textbf{FPS}}} \\
        \cline{2-8}
            & 
            \multirow{1.3}*{PR} & 
             \multirow{1.3}*{SR} & 
             \multirow{1.3}*{PR} & 
             \multirow{1.3}*{RE} & 
             \multirow{1.3}*{F-score} & 
             \multirow{1.3}*{PR} & 
             \multirow{1.3}*{SR} & 
             & \\
        \midrule
        Crs\_Attn & 67.6 & 54.4 & 61.2 & 61.4 & 61.3 & 75.8 & 59.3 & 56.4 & 41.5 \\
        \rowcolor{cyan!5}+AMG-LoRA & 69.9 & 55.7 & 62.7 & 60.4 & 61.5 & 76.4 & 59.9 & 56.4 & 41.2 \\
        \midrule
        MCP & 68.4 & 54.9 & 60.1 & 59.2 & 59.6 & 75.6 & 59.3 & 56.5 & 54.7 \\
        \rowcolor{cyan!5}+AMG-LoRA & 70.3 & 56.1 & 61.2 & 60.7 & 61.0 & 75.9 & 59.5 & 56.5 & 54.2 \\
        \midrule
        HMoE & 67.4 & 54.2 & 60.9 & 61.3 & 61.1 & 76.5 & 59.9 & 56.4 & 63.8 \\
        \rowcolor{cyan!15}+AMG-LoRA & \textbf{71.6} & \textbf{57.3} & \textbf{62.9} & \textbf{63.5} & \textbf{63.2} & \textbf{77.1} & \textbf{60.3} & \textbf{56.4} & \textbf{63.5}\\
        \bottomrule
    \end{tabular}
    }
    \vspace{-17pt}
\end{table}
\paragraph{Visualization Analysis.}AMG-LoRA is designed to handle modality reliability variations in tracking. Although $w_{\text{X}}$ (Eq.~\ref{eq:dte2r}) and $w_{\text{rgb}}$ (Eq.~\ref{eq:r2dte}) remain fixed during inference, the results exhibit interesting adaptivity. Fig.~\ref{fig:vis_amglora} provides intuitive evidence of its effect, where the ``Pre-train” row denotes attention maps directly inferred from the foundation tracker. Rather than being misled by the less reliable modality (\textit{i.e.}, negative transfer), AMG-LoRA facilitates mutual refinement, producing clean and well-aligned matching responses in the RGB-T task (Case 1). Cases 2 and 3 further demonstrate that such alignment is conditioned on the correct branch. Fig.~\ref{fig:vis_bbox} further provides a prediction-level comparison with well-established trackers. All three cases involve similar-appearance scenarios, a common challenge in real-world applications. Whereas existing methods are easily misled and yield incorrect predictions, SEATrack remains accurate across all cases, which we attribute to the “denoising” capability of AMG-LoRA.
\vspace{-5pt}
     

\section{Conclusion}
In this work, we revisit the limitations of existing PEFT multimodal tracking methods and present SEATrack to address the performance–efficiency dilemma. Rather than focusing solely on cross-modal fusion, it adopts an alignment-before-fusion design. Extensive evaluations across multiple tasks demonstrate that this principle contributes to the simplicity and effectiveness of SEATrack.

\textbf{Limitation.} While effective, SEATrack underperforms on certain metrics. The input-dependent alignment mechanisms may further improve robustness. Moreover, exploring how to align spatially heterogeneous modalities, such as vision and language, in the tracking context remains an interesting direction for future work.

\paragraph{Acknowledgement.}The authors thank the reviewers and ACs for their tremendous efforts and helpful comments. This work is partially supported by the Natural Science Foundation of China (No. 62476235 and No. 62503323); the Hebei Natural Science Foundation (No. F2023203012); and the Innovation Capability Improvement Plan Project of Hebei Province (No. 22567626H).
{
    \small
    \bibliographystyle{ieeenat_fullname}
    \bibliography{main}
}
\clearpage
\setcounter{page}{1}
\maketitlesupplementary

\section{Overall}
\label{sec:overall}
In this supplementary material, we provide more exploration and analyses of the proposed Adaptive Mutual Guidance Low-Rank Adaptation (AMG-LoRA) and Hierarchical Mixture of Experts (HMoE), which are difficult to describe in the main paper due to space limitations. Specifically, the content of the supplementary material is organized below:
\begin{itemize}
\item The rank choices of AMG-LoRA and HMoE;
\item Types of Adaptation Weights in AMG-LoRA;
\item Statistical Results of Alignment;
\item More Visualizations;
\item Complexity Analysis of HMoE;
\item Mechanism Behind Adaptive Alignment.
\end{itemize}

\subsection{Rank Choices of AMG-LoRA and HMoE}
Low-rank approximation facilitates the parameter efficiency of AMG-LoRA and HMoE. \cref{tab:rank_choices} systematically ablates different rank choices within each component to identify the most effective configuration.
\paragraph{AMG-LoRA:}\cref{tab:amglora-rank}(a) shows that setting $r=8$ for AMG-LoRA achieves the best performance across all three tasks, which is consistent with the findings of~\cite{hu2022lora}. In contrast, smaller or larger rank choices lead to under-adaptation and over-adaptation of the pretrained knowledge, respectively, resulting in suboptimal performance.

\begin{table}[h!]
    \centering
    \caption{Rank Choices of AMG-LoRA and HMoE.}
    \label{tab:amglora-rank}
    \begin{subtable}[h!]{0.8\linewidth}
        \caption{AMG-LoRA}
        \resizebox{\linewidth}{!}{
        \begin{tabular}{c|cc|ccc|cc|c}
            \toprule
            \multicolumn{1}{c|}{\multirow{2}{*}{$r$}} &
            \multicolumn{2}{c|}{\textbf{LasHeR}} & 
            \multicolumn{3}{c|}{\textbf{DepthTrack}} & 
            \multicolumn{2}{c|}{\textbf{VisEvent}} &
            \multicolumn{1}{c}{\multirow{2}{*}{\textbf{Param.}}}\\
            \cline{2-8}
                & 
                \multirow{1.3}*{PR} & 
                \multirow{1.3}*{SR} & 
                \multirow{1.3}*{PR} & 
                \multirow{1.3}*{RE} & 
                \multirow{1.3}*{F-score} & 
                \multirow{1.3}*{PR} & 
                \multirow{1.3}*{SR} &
                \\
            \midrule
            4 & 69.9 & 56.0 & 58.7 & 59.2 & 59.0 & 76.2 & 59.7 & 0.5M\\
            \midrule
            \rowcolor{cyan!15} \textbf{8} & \textbf{71.6} & \textbf{57.3} & \textbf{62.9} & \textbf{63.5} & \textbf{63.2} & \textbf{77.1} & \textbf{60.3} & \textbf{0.6M}\\
            \midrule
            16 & 70.3 & 56.1 & 60.4 & 60.8 & 60.6 & 76.7 & 60.2 & 0.8M\\
            \bottomrule
        \end{tabular}
        }
    \end{subtable}
    \begin{subtable}[h!]{0.8\linewidth}
        \caption{HMoE}
        \label{tab:hmoe-rank}
        \resizebox{\linewidth}{!}{
        \begin{tabular}{c|cc|ccc|cc|c}
            \toprule
            \multicolumn{1}{c|}{\multirow{2}{*}{$r$}} &
            \multicolumn{2}{c|}{\textbf{LasHeR}} & 
            \multicolumn{3}{c|}{\textbf{DepthTrack}} & 
            \multicolumn{2}{c|}{\textbf{VisEvent}} &
            \multicolumn{1}{c}{\multirow{2}{*}{\textbf{Param.}}}\\
            \cline{2-8}
                & 
                \multirow{1.3}*{PR} & 
                \multirow{1.3}*{SR} & 
                \multirow{1.3}*{PR} & 
                \multirow{1.3}*{RE} & 
                \multirow{1.3}*{F-score} & 
                \multirow{1.3}*{PR} & 
                \multirow{1.3}*{SR} &\\
            \midrule
            \rowcolor{cyan!15} \textbf{4} & \textbf{71.6} & \textbf{57.3} & \textbf{62.9} & \textbf{63.5} & \textbf{63.2} & \textbf{77.1} & \textbf{60.3} & \textbf{0.6M}\\
            \midrule
            8 & 69.7 & 55.9 & 60.4 & 59.5 & 60.0 & 76.2 & 59.9 & 1M\\
            \midrule
            16 & 69.0 & 55.2 & 58.2 & 58.2 & 58.2 & 76.2 & 59.8 & 1.7M\\
            \bottomrule
        \end{tabular}
        }
    \end{subtable}
    \label{tab:rank_choices}
\end{table}

\paragraph{HMoE:}Unlike AMG-LoRA, which focuses on adapting pretrained knowledge, the low-rank approximation in HMoE aims to learn effective multimodal relation modeling in a parameter-efficient manner. \cref{tab:amglora-rank}(b) indicates that $r=4$ is sufficient given the current data size (0.51M, 0.22M, and 0.21M frame pairs for LasHeR~\cite{li2021lasher}, DepthTrack~\cite{yan2021depthtrack}, and VisEvent~\cite{wang2023visevent}, respectively). Increasing the rank to further expand the model capacity instead leads to degraded performance, likely due to overfitting.

\begin{table}[h!]
    \caption{Types of Adaptation Weights in AMG-LoRA}
    \label{tab:adaptation_weight}
    \centering
    \resizebox{0.8\linewidth}{!}{
    \begin{tabular}{c|cc|ccc|cc}
        \toprule
        \multicolumn{1}{c|}{\multirow{2}{*}{Weight Type}} &
        \multicolumn{2}{c|}{\textbf{LasHeR}} & 
        \multicolumn{3}{c|}{\textbf{DepthTrack}} & 
        \multicolumn{2}{c}{\textbf{VisEvent}} \\
        \cline{2-8}
            & 
            \multirow{1.3}*{PR} & 
            \multirow{1.3}*{SR} & 
            \multirow{1.3}*{PR} & 
            \multirow{1.3}*{RE} & 
            \multirow{1.3}*{F-score} & 
            \multirow{1.3}*{PR} & 
            \multirow{1.3}*{SR}\\
        \midrule
        $W_{[q,k]}$ & 69.8 & 56.1 & 56.8 & 56.4 & 56.6 & 75.8 & 59.4 \\
        \midrule
        $W_{[q,v]}$ & 71.3 & 56.8 & 60.3 & 59.6 & 60.0 & 76.0 & 59.4 \\
        \midrule
        \rowcolor{cyan!15} \textbf{$W_{[k,v]}$} & \textbf{71.6} & \textbf{57.3} & \textbf{62.9} & \textbf{63.5} & \textbf{63.2} & \textbf{77.1} & \textbf{60.3} \\
        \bottomrule
    \end{tabular}
    }
\end{table}

\subsection{Types of Adaptation Weights in AMG-LoRA}
In \cref{subsec:amglora}, we leverage LoRA~\cite{hu2022lora} to adapt the $[W_k,W_v]$ in the attention layer, which differs from the original setting that adapts $[W_q,W_v]$. \cref{tab:adaptation_weight} provides empirical evidence supporting this choice. It can be inferred that adapting $W_v$ is crucial for cross-modal alignment, and combining it with $W_k$ instead of $W_q$ yields better performance and generalization across the three tasks.

\begin{table}[h!]
    \caption{Statistical results of alignment. Cos denotes cosine similarity, and SKL denotes symmetric KL divergence (scaled by 10\textsuperscript{4} for clarity).}
    \label{tab:alignment}
    \centering
    \resizebox{1\linewidth}{!}{
    \begin{tabular}{c|cc|cc|cc}
        \toprule
        \multicolumn{1}{c|}{\multirow{2}{*}{Method}} &
        \multicolumn{2}{c|}{\textbf{LasHeR}} & 
        \multicolumn{2}{c|}{\textbf{DepthTrack}} & 
        \multicolumn{2}{c}{\textbf{VisEvent}} \\
        \cline{2-7}
            & 
            \multirow{1.3}*{Cos($\uparrow$)} & 
            \multirow{1.3}*{SKL($\downarrow$)} & 
            \multirow{1.3}*{Cos($\uparrow$)} & 
            \multirow{1.3}*{SKL($\downarrow$)} & 
            \multirow{1.3}*{Cos($\uparrow$)} & 
            \multirow{1.3}*{SKL($\downarrow$)}\\
        \midrule
        LoRA & 0.51 & 0.99 & 0.44 & 1.2 & 0.51 & 0.92 \\
        \midrule
        \rowcolor{cyan!15} \textbf{AMG-LoRA} & \textbf{0.99} & \textbf{0.04} & \textbf{0.97} & \textbf{0.09} & \textbf{0.91} & \textbf{0.16} \\
        \bottomrule
    \end{tabular}
    }
\end{table}

\subsection{Statistical Results of Alignment}
\cref{fig:amglora_vs_lora} demonstrates the benefits of alignment for multimodal tracking by showing AMG-LoRA's performance gains over LoRA across challenging attributes. While \cref{fig:modality-missing_vis} and \cref{fig:vis_amglora} provide intuitive qualitative evidence of AMG-LoRA’s adaptive refinement capability. Here, we provide a purely quantitative assessment of the alignment improvements achieved by AMG-LoRA.

Specifically, we randomly sample 300 frames per task for inference and measure alignment by computing the similarity between the attention maps (matching responses). With cosine similarity for unnormalized attention maps (pre-softmax) and symmetric KL divergence for normalized attention maps (post-softmax).

We aggregate the layer-wise results and report their average in \cref{tab:alignment}. AMG-LoRA clearly drives the attention maps across modalities toward near identity, strongly validating the motivation for alignment.

\begin{figure*}[t!]
    \centering
    \includegraphics[width=\linewidth]{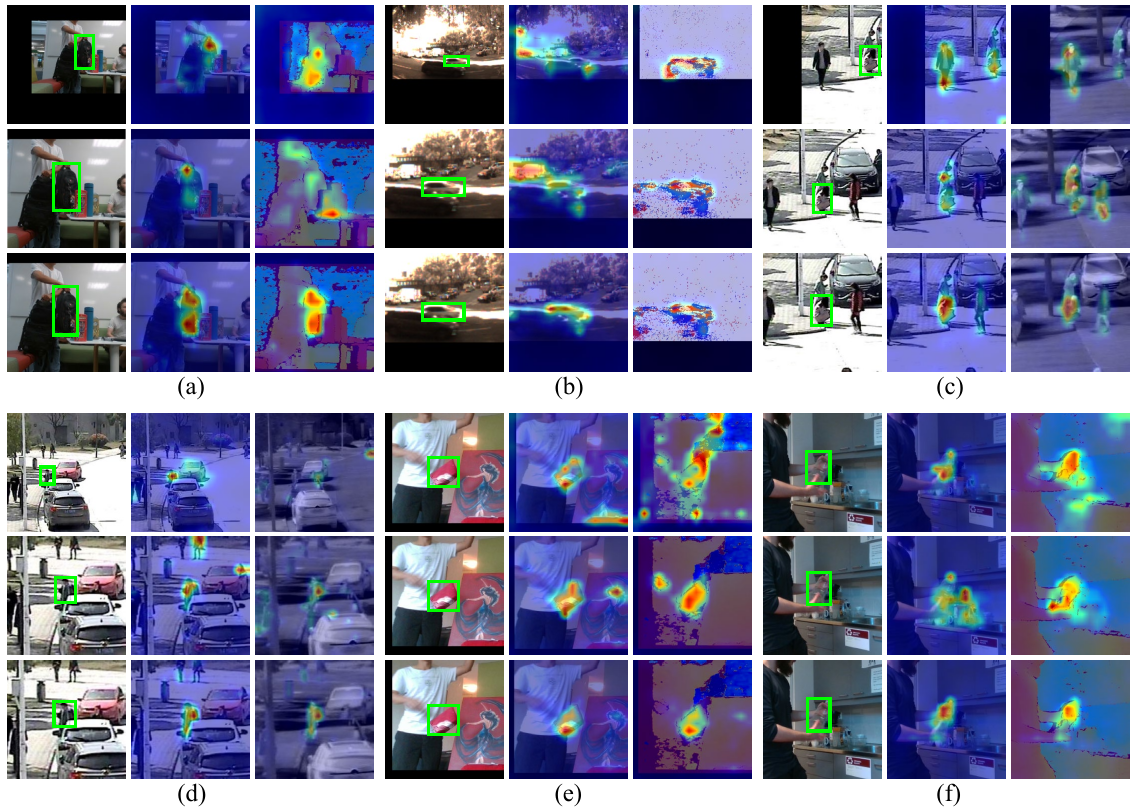}
    \caption{Additional visualizations of the adaptability of AMG-LoRA under diverse real-world scenarios. For each case, the columns present the ground-truth target region (\textbf{left}), the corresponding RGB branch attention map (\textbf{middle}), and the corresponding X-modal attention map (\textbf{right}). While the rows correspond to the outputs of the Foundation model~\cite{ye2022joint}, LoRA~\cite{hu2022lora}, and the proposed AMG-LoRA.
    }
    \label{fig:vis_amglora+}
\end{figure*}

\subsection{More Visualization}
\cref{fig:vis_amglora+} provides additional cases, including both indoor and outdoor real-world scenarios, further illustrating the adaptability of AMG-LoRA. By dynamically refining the less reliable matching response (attention map) instead of negative transferring, it demonstrates strong potential for all-time, all-weather tracking.

\subsection{Complexity Analysis of HMoE}
For HMoE, the main computational overhead comes from Eqs.~\ref{eq:6},~\ref{eq:7},~\ref{eq:9},~\ref{eq:10}; with the softmax in Eq.~\ref{eq:6} being the dominant term at $O(Neh^{2})$. Since the number of heads $h$ is a small constant and much smaller than the number of tokens $N$, \textit{i.e.} $h\ll N$, HMoE is therefore more efficient than attention with $O(N^{2}D)$ complexity.

\begin{table}[h!]
    \centering
    \caption{Efficiency under Different HMoE Configurations.}
    \label{tab:amglora-rank}
    \hfill
    \begin{subtable}[h!]{0.4\linewidth}
        \caption{Heads}
        \resizebox{\linewidth}{!}{
        \begin{tabular}{c|c|c}
            \toprule
            \textbf{Heads} &
            \textbf{MACs ($10^9$)} &
            \textbf{FPS} \\
            \midrule
            1 & 56.466 & 63.7 \\
           \rowcolor{cyan!15} 2 & 56.466 & 63.5 \\
            4 & 56.466 & 62.4\\
            8 & 56.466 & 62.0 \\
            \bottomrule
        \end{tabular}
        }
    \end{subtable}
    \hfill
    \begin{subtable}[h!]{0.47\linewidth}
        \caption{Experts}
        \resizebox{\linewidth}{!}{
        \begin{tabular}{c|c|c|c}
            \toprule
            \textbf{Attn.} &
            \textbf{FFN.} &
            \textbf{MACs ($10^9$)} &
            \textbf{FPS} \\
            \midrule
            4 & 4 & 56.466 & 64.1 \\
            8 & 4 & 56.466 & 63.5 \\
            \rowcolor{cyan!15} 4 & 8 & 56.466 & 63.5 \\
            8 & 16 & 56.466 & 62.4 \\
            \bottomrule
        \end{tabular}
        }
    \end{subtable}
    \hfill
    \label{tab:10}
\end{table}

Following the configurations in Tab.\ref{tab:head_number_HMoE} and Tab.~\ref{tab:expert_number_HMoE}, we further investigate the efficiency of HMoE under varying numbers of heads and experts, where MACs are measured using the THOP package. As shown in Tab.~\ref{tab:10}(a), increasing the number of heads $h$ from 1 to 8 introduces only negligible computational overhead at the G-level, while the FPS variations remain within the margin of error. Tab.\ref{tab:10}(b) further shows that increasing the number of experts $e$ is also largely insensitive to efficiency, which makes it possible to scale the model without introducing additional latency. These observations are consistent with the theoretical complexity analysis mentioned above.

\begin{figure}[h!]
    \centering
    \includegraphics[width=\linewidth]{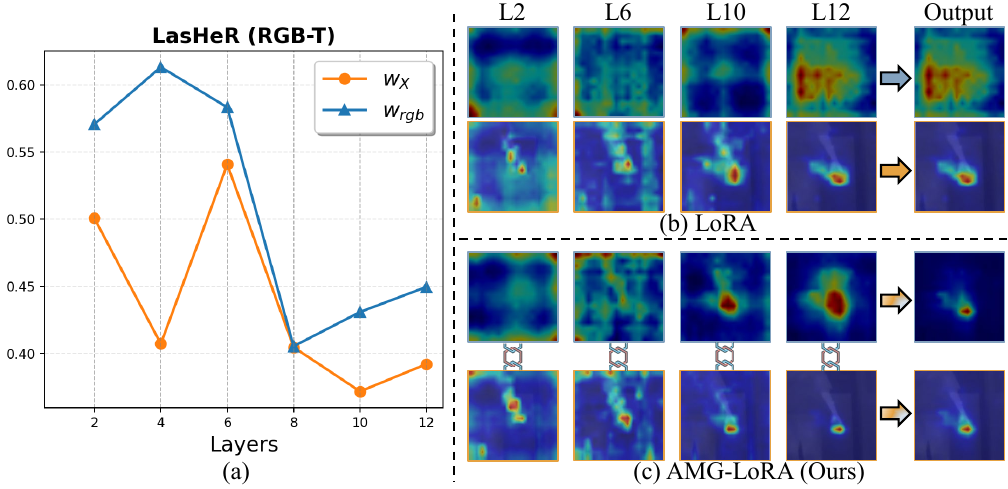}
    \caption{(a): the learned scalars across layers. (b) and (c): layer-wise attention maps (top: RGB; bottom: Thermal).
    }
    \label{fig:re}
\end{figure}

\subsection{Mechanism Behind Adaptive Alignment}
To investigate the mechanism behind the adaptive alignment behavior of AMG-LoRA shown in Fig.~\ref{fig:modality-missing_vis} and Fig.~\ref{fig:vis_amglora}, we first analyze the layer-wise distributions of the two learnable scaling factors, $w_{\mathrm{rgb}}$ and $w_\text{X}$, learned on the LasHeR dataset. As illustrated in Fig.~\ref{fig:re}(a), the guidance strength of $w_{\mathrm{rgb}}$ is higher than that of $w_\text{X}$ in most layers, indicating that, in RGB-T scenarios, the RGB modality generally provides more reliable target responses. Meanwhile, $w_\text{X}$ remains positive throughout all layers, suggesting that the thermal branch also contributes useful complementary information during alignment. This indicates that the proposed mutual guidance mechanism does not rigidly favor one branch; instead, it learns a collaborative pattern during training, in which the more reliable branch takes the leading role while the other serves as complementary support.

To better understand the behavior of AMG-LoRA under frame loss, we take frame \textit{\textbf{\#46}} in Fig.~\ref{fig:modality-missing_vis} as an example and compare its layer-wise attention maps with those of the baseline (LoRA), as shown in Fig.~\ref{fig:re}(b) and (c). In contrast to the consistently dispersed patterns produced by LoRA, our AMG-LoRA enables the impaired RGB branch to gradually refine its target perception in deeper layers, thereby laying the foundation for accurate final alignment. We attribute this behavior to the effective gradients introduced by the proposed Adaptive Mutual Guidance (AMG), which help the pretrained knowledge learn how to align correctly during training.


\end{document}